\documentclass[10pt,twocolumn,letterpaper]{article}

\usepackage[pagenumbers]{cvpr} % To force page numbers, e.g. for an arXiv version

\usepackage{graphicx}
\usepackage{comment}
\usepackage{amsmath,amssymb} % define this before the line numbering.
\usepackage{color}
\usepackage[export]{adjustbox}
\usepackage[inline]{enumitem}
\usepackage{booktabs}
\usepackage{array,multirow,graphicx}
\usepackage{siunitx}
\usepackage{pifont}
\usepackage{textcomp}

\usepackage[pagebackref,breaklinks,colorlinks]{hyperref}

\usepackage[accsupp]{axessibility}

\usepackage[capitalize]{cleveref}
\crefname{section}{Sec.}{Secs.}
\Crefname{section}{Section}{Sections}
\Crefname{table}{Table}{Tables}
\crefname{table}{Tab.}{Tabs.}

\def\cvprPaperID{1783} % *** Enter the CVPR Paper ID here
\def\confName{CVPR}
\def\confYear{2022}

\begin{document}

\title{REX: Reasoning-aware and Grounded Explanation}

\author{Shi Chen \qquad Qi Zhao\\
Department of Computer Science and Engineering,\\
University of Minnesota\\
{\tt\small \{chen4595, qzhao\}@umn.edu}}

\maketitle

%%%%%%%%% ABSTRACT
\begin{abstract}
Effectiveness and interpretability are two essential properties for trustworthy AI systems. Most recent studies in visual reasoning are dedicated to improving the accuracy of predicted answers, and less attention is paid to explaining the rationales behind the decisions. As a result, they commonly take advantage of spurious biases instead of actually reasoning on the visual-textual data, and have yet developed the capability to explain their decision making by considering key information from both modalities. This paper aims to close the gap from three distinct perspectives: first, we define a new type of multi-modal explanations that explain the decisions by progressively traversing the reasoning process and grounding keywords in the images. We develop a functional program to sequentially execute different reasoning steps and construct a new dataset with 1,040,830 multi-modal explanations. Second, we identify the critical need to tightly couple important components across the visual and textual modalities for explaining the decisions, and propose a novel explanation generation method that explicitly models the pairwise correspondence between words and regions of interest. It improves the visual grounding capability by a considerable margin, resulting in enhanced interpretability and reasoning performance. Finally, with our new data and method, we perform extensive analyses to study the effectiveness of our explanation under different settings, including multi-task learning and transfer learning. Our code and data are available at \url{https://github.com/szzexpoi/rex}.

\end{abstract}

%%%%%%%%% BODY TEXT
\section{Introduction}
\begin{figure}
\centering
\includegraphics[width=1\linewidth]{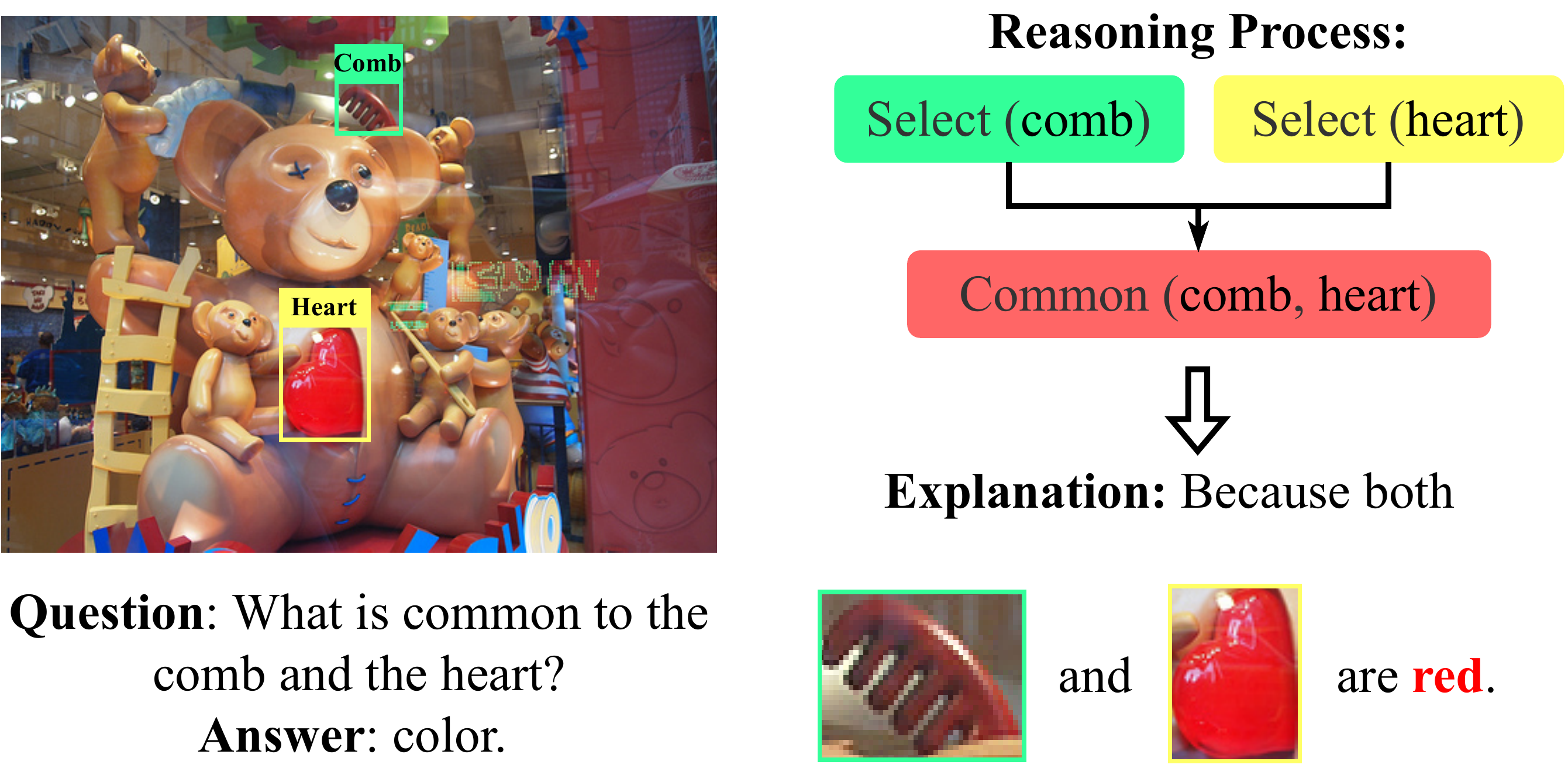}
\caption{Illustration of our explanation that is derived from the reasoning process (with different reasoning steps color coded) and explicitly grounds key objects in the image. %Besides explaining the answers, it also enables the analyses of various visual skills.
}
\label{fig:teaser}
\end{figure}

One of the fundamental goals in artificial intelligence is to develop intelligent systems that are able to reason and explain with the complexity of real-world data to make decisions. While explaining decisions is an integral part of human communication, understanding and reasoning, existing visual reasoning models typically answer questions without explaining the rationales behind their answers. As a result, despite the significantly increased accuracy achieved by powerful deep neural networks \cite{updown,ban,lxmert,vilbert,visualbert,oscar}, existing methods commonly take advantage of spurious data biases \cite{explicit_bias} and it is difficult to understand if they make decisions by truly understanding the causal relationships between multi-modal inputs and the answers.
% it is still relatively prohibitive to understand if a model makes decisions by truly understanding the causal relationships between multi-modal inputs and the answer or simply taking advantage of spurious data biases \cite{explicit_bias}.  

An important line of research to tackle the issues is to improve the interpretability of visual reasoning models with multi-modal explanations \cite{vqa_x,vqa_e,vcr,faithful_exp,transformer_exp,generative_vcr,competing_exp}. While showing usefulness in highlighting important visual regions and providing user-friendly textual descriptions, these approaches suffer from two major limitations: (1) Existing explanations are typically defined in the forms of attention maps or free-formed natural language. Attention maps capture the salient regions for generating the answers but fall short of explaining how different regions contribute to the decision-making process. On the other hand, unconstrained textual explanations could be highly diverse and often inconsistent when explaining the same decision. Both of them lack the capability to illustrate the reasoning process behind a decision. (2) The explanations of different modalities are loosely connected and modeled with separate processes \cite{vqa_x,vqa_e,faithful_exp}. This not only undermines the capability of explaining models' rationales with multiple modalities, but can also result in contradictory explanations \cite{faithful_exp}. For instance, textual explanations ``The apple is above the pear'' and ``The pear is above the apple'' have opposite meanings but could share the same attention map. We address the aforementioned challenges from two distinct perspectives (\textit{i.e.,} data and model), and propose an integrated framework that consists of a new type of explanations, a functional program, and a novel explanation generation method. 

From the data perspective, instead of independently modeling explanations of a single modality without considering the reasoning process, we introduce a new Reasoning-aware and grounded EXplanation (REX) that is derived by traversing the reasoning process and tightly coupling key components across the visual and textual modalities. As shown in Figure \ref{fig:teaser}, it is constructed based on the consecutive reasoning steps (\eg, select, common) for decision making, and explicitly grounds key objects (\eg, comb, heart) with visual regions to elaborate how they contribute to the answer. The structured reasoning process also naturally alleviates the variance of natural language, and enables models to pay focused attention to important information for reasoning instead of the language structure. To automatically construct our explanations, we develop a functional program to progressively execute the reasoning steps and query key information from scene graphs \cite{scene_graph,visual_genome}, and collect a new dataset with 1,040,830 multi-modal explanations. 

% The structured reasoning process also naturally alleviates the significant variance of natural language, while the explicit grounding allows us to model the pairwise relationships between contents of the two modalities and how they correlate with the answer. To automatically construct our defined explanation, we develop a functional program to progressively execute the reasoning steps, and collect a new dataset with 1,040,830 explanations. 

% From the data perspective, instead of independently modeling explanations of a single modality without considering the reasoning process, we introduce a new REasoning-aware and Grounded EXplanation (REX). As shown in Figure \ref{fig:teaser}, our explanation is derived from a set of consecutive reasoning steps and tightly couples key components across the visual and textual modalities.
% Through traversing the reasoning process to gather information from visually grounded objects, it explains how decisions are made with the contents from the two modalities, highlighting the pairwise relationships and correlation with the answers. The structured reasoning process also naturally alleviates the significant variance of natural language. % and enables the analyses of essential visual skills for tackling the problems (\textit{e.g.,} visual grounding and recognition of colors). 
% To automatically construct our defined explanation, we develop a functional program to progressively execute the reasoning steps, and collect a new dataset with 1,040,830 explanations. 

From the model perspective, unlike existing methods \cite{vqa_x,vqa_e,faithful_exp,transformer_exp,generative_vcr} that model key components in different modalities with separate processes, we propose a novel explanation generation method that explicitly models the correspondence between important words and regions of interest. It takes into account the semantic similarity between features of the two modalities, and incorporates an adaptive gate to ground words in the visual scene. Our method improves the visual grounding by a large margin, resulting in enhanced interpretability and reasoning performance.

% To summarize, our contributions are as follows:
% \begin{itemize}
%     \item We present REX, a new type of reasoning-aware and visually grounded explanation. Our explanation differentiates itself with its strong correlation with the reasoning process and the tight coupling between different modalities. We develop a functional program to automatically construct our new dataset with 1,040,830 multi-modal explanations.
%     \item We propose a novel explanation generation method that goes beyond the conventional paradigm of independently modeling multi-modal explanations \cite{faithful_exp,transformer_exp,generative_vcr}, and leverages an explicit mapping to ground words in the visual regions based on their correlation.
%     \item We demonstrate the effectiveness of our data and method with extensive experiments under different settings, including multi-task learning and transfer learning. We also analyze different visual skills and their correlation with the reasoning performance.
% \end{itemize}

To summarize, our contributions are as follows:

(1) We present REX, a new type of reasoning-aware and visually grounded explanations. Our explanation differentiates itself with its strong correlation with the reasoning process and the tight coupling between different modalities. We develop a functional program to automatically construct our new dataset with 1,040,830 multi-modal explanations.

(2) We propose a novel explanation generation method that goes beyond the conventional paradigm of independently modeling multi-modal explanations \cite{faithful_exp,transformer_exp,generative_vcr}, and leverages an explicit mapping to ground words in the visual regions based on their correlation.

(3) We demonstrate the effectiveness of our data and method with extensive experiments under different settings, including multi-task learning and transfer learning. We also analyze different visual skills and their correlation with the reasoning performance.

\section{Related Works}
This paper is related to previous efforts on visual question answering (VQA), multi-modal explanation datasets for visual reasoning, and explanation generation models.

\textbf{Visual question answering.} Visual reasoning is commonly framed as a VQA task. There is a large body of research on constructing VQA datasets \cite{vqa1,vqa2,gqa,vcr,clevr,scene_text_qa,visualcomet,okvqa} and developing VQA models \cite{mcb,mfb,mlb,updown,ban,stacknmn,xnm,nsm,lxmert,vilbert,visualbert}. Early VQA datasets typically collect human-annotated questions through crowd-sourcing \cite{vqa1,vqa2,vcr}. Several recent studies \cite{gqa,clevr} propose to use functional programs to automatically generate questions based on pre-defined rules and enable more balanced distributions of question-answer pairs. There is also an increasing interest in investigating different types of visual reasoning, \textit{e.g.,} scene text understanding \cite{scene_text_qa}, reasoning on dynamic context \cite{visualcomet}, and knowledge-based reasoning \cite{okvqa}. These data efforts lead to the development of computational approaches for improving different components of VQA models, including multi-modal fusion \cite{mcb,mfb,mlb}, attention mechanism \cite{updown,ban}, and inference process \cite{stacknmn,xnm,nsm}. Vision-and-language pretraining \cite{lxmert,vilbert,visualbert,oscar} has also shown usefulness in VQA for enhancing the understanding of multi-modality. Our framework is complementary to existing efforts in VQA. It augments existing VQA data with reasoning-aware and visually grounded explanations, and enables the development of VQA models with enhanced interpretability and reasoning performance.

\begin{figure*}
\centering
\includegraphics[width=0.8\linewidth]{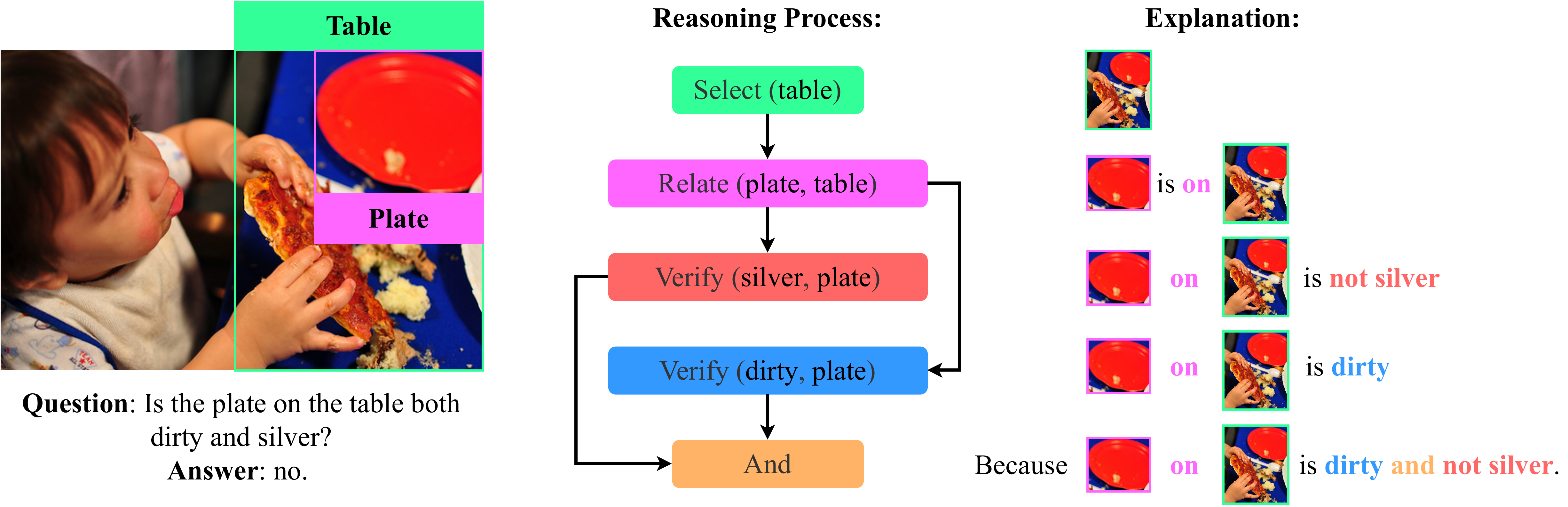}
\caption{Illustration of the process for sequentially constructing our explanation. Partial explanations are shown to the right of the corresponding reasoning steps, and information collected from different steps is highlighted with their corresponding colors. The final explanation is obtained at the end of the reasoning process.}
\label{fig:data}
\end{figure*}

\textbf{Multi-modal explanations for visual reasoning.} There is a dearth of studies that construct multi-modal explanation datasets for visual reasoning. The pioneering work \cite{vqa_x} collects 41,817 textual explanations annotated by humans on the VQA datasets \cite{vqa1,vqa2}, and develops a visual pointing task to highlight important regions. To automatically build large-scale explanation datasets, Li \textit{et al.} \cite{vqa_e} propose to convert captioning annotations \cite{mscoco} to textual explanations. By estimating the similarity between captions and questions, it generates 269,786 synthetic explanations. Zellers \textit{et al.} \cite{vcr} propose a multi-choice VQA dataset with 264,720 questions and each question is associated with a correct explanation. %While the work also considers visual grounding for explanation, as their dataset focuses on movie scenes, about 91$\%$ of the grounded objects are humans and over 40$\%$ of the explanations do not have visual grounding. 
While the work has annotated regions of interest for the questions, as their dataset focuses on movie scenes, about 91$\%$ of the regions are related to human characters and over 40$\%$ of the explanations can not be grounded on them. The key differentiators of our explanation lie in its correlation with the reasoning process for problem-solving and the explicit coupling between the key components in different modalities. Our dataset offers 1,040,830 structured and visually grounded explanations that are aware of the decision-making process and ground various keywords in the images. % and enables the analyses of different visual skills for reasoning and their relationships with reasoning performance. % It opens a new avenue for studying the underlying rationales behind visual reasoning models with multi-modal explanations.

\textbf{Generating multi-modal explanations.} Aiming to develop visual reasoning models that are capable of explaining their answers, several works propose to automatically generate multi-modal explanations. Park \textit{et al.} \cite{vqa_x} use a long short-term memory (LSTM) model to generate the textual explanations, and highlight important visual evidence with attention maps. Later on, Wu \textit{et al.} \cite{faithful_exp} improve the model by correlating the attention for answering questions and generating explanations. Li \textit{et al.} \cite{vqa_e} propose a multi-task learning paradigm to simultaneously generate answers and explanations. Instead of predicting explanations from scratch, Zellers \textit{et al.} \cite{vcr} adopt a multi-choice task setting whose goal is to select the correct explanation out of four candidates. Marasovi{\'c} \textit{et al.} \cite{transformer_exp} develop an integrated method that incorporates pretrained language models with object recognition models. Dua \textit{et al.} \cite{generative_vcr} frame both VQA and explanation generation as generative tasks, and sequentially generates words in answers and explanations. There are also studies \cite{competing_exp,self_critical_vqa} that leverage generated or ground truth multi-modal explanations to improve the reasoning performance. Different from the aforementioned methods that independently model visual or textual explanations, our proposed method explicitly links words with the corresponding image regions based on their semantic similarity. The enhanced visual grounding capability brought by our method not only improves the interpretability and reasoning performance, but also plays a key role in distilling knowledge from explanations into question answering.

% Unlike the aforementioned methods that model explanations of different modalities with separate process, our proposed explanation generation method explicitly grounds keywords in textual explanation in the corresponding image regions and tightly couples different modalities. 

\section{Reasoning-aware and Grounded Explanation}
Answering visual questions would benefit from capabilities of reasoning on multi-modal contents and explaining the answers. This section presents a principled framework for visual reasoning with enhanced interpretability and effectiveness. It advances the research in visual reasoning from both the data and the model perspectives with:
\begin{enumerate*}
    \item[(1)] a new type of multi-modal explanations that explain decision making by traversing the reasoning process, together with a functional program to automatically construct the explanations, and
    \item[(2)] an explanation generation method that explicitly models the relationships between words and visual regions, and simultaneously enhances the interpretability as well as reasoning performance. 
\end{enumerate*}

\subsection{Data}
The goal of our proposed data is to offer an explanation benchmark that encodes the reasoning process and grounding across the visual-textual modalities. Compared to previous explanations for visual reasoning \cite{vqa_x,vqa_e,vcr}, it has two key advantages: (1) Grounded on the reasoning process, it elaborates how different components in the visual and textual modalities contribute to the decision making, and reduces variance or inconsistency in textual descriptions; and (2) Instead of modeling textual and visual explanations as separate components, our explanation considers evidence from both modalities in an integral manner, and tightly couples words with image regions (\textit{i.e.,} for visual objects, their grounded regions instead of object names are considered in the explanations). It augments visual reasoning models with the capability to explain their decision making by jointly considering both modalities, resulting in enhanced interpretability and reasoning performance.

% It not only provides a more comprehensive illustration of the decision-making process, but also augments the visual reasoning models with enhanced visual grounding capability and reasoning performance.  

% We describe the reasoning process as a sequence of atomic operations, and derive our explanations by extracting multi-modal information with each operation (\textit{e.g.,} visually grounded objects and their attributes) and accumulating information across different operations based on their semantic meanings (\textit{e.g.,} the \textit{And} operation performs a logical ``and'' on the objects selected in the previous operations). 

Figure \ref{fig:data} illustrates the paradigm for constructing our explanation. To answer the question ``Is the plate on the table both dirty and silver?'', one needs to locate the table, find the plate on top of it based on their relationship, and investigate the cleanliness as well as the color of the plate. We represent each reasoning step with an atomic operation, \textit{e.g.,} \textit{select} and \textit{verify}, and leverage a functional program to sequentially construct the explanation by traversing the reasoning steps and accumulating important information (\textit{e.g.,} visually grounded objects and their attributes). Upon finishing the traversal, our final explanation not only elaborates the decision making with concrete textual description (\textit{i.e.,} the plate is dirty but not silver thus the answer is no), but also supports the explanation with visual evidence (\textit{i.e.,} grounded regions for the plate and the table).

\begin{table}[t]
\begin{center}
\resizebox{0.9\linewidth}{!}{
% {\scriptsize
\begin{tabular}{cc}
\toprule
\textbf{Operation} & \textbf{Semantic}  \\
\midrule
Select & Selecting a specific category of objects. \\
\midrule
Exist & Examining the existence of a specific type of objects. \\
\midrule
Filter & Selecting the targeted objects by looking for a specific attribute.  \\
\midrule
Query & Retrieving the value of a attribute from the selected objects.  \\
\midrule
Verify & Examining if the targeted objects have a given attribute.\\
\midrule
Common & Finding the common attributes among a set of objects. \\
\midrule
Same & Examining if two groups of objects have the same attribute.\\
\midrule
Different & Examining if two groups of objects have different attributes.\\
\midrule 
Compare & Comparing the values of an attribute between multiple objects.\\
\midrule
Relate & Connecting different objects using their relationships. \\
\midrule
And/Or & Logical operations that combine results of previous operations.\\
\bottomrule
\end{tabular}
}
\end{center}
\caption{Atomic operations to represent the reasoning process.}
\label{atomic_operation}
\end{table}

\textbf{Decomposing the reasoning process with atomic operations.} We define a vocabulary of atomic operations by characterizing and abstracting functions for question generation in the GQA dataset \cite{gqa}. Given the 127 different types of operations in GQA, we first follow \cite{air} and represent each operation as a triplet, \textit{i.e.,} $<$operation, attribute, category$>$, and then categorize the original operations in GQA programs based on their semantic meanings. As shown in Table \ref{atomic_operation}, we define 12 atomic operations that cover the essential steps for answering various types of visual questions: some require localizing a specific type of objects (\textit{select}, \textit{exist}); some require reasoning on attributes of the objects (\textit{filter}, \textit{query}, \textit{verify}, \textit{common}, \textit{same}, \textit{different}, \textit{compare}, \textit{relate}); and others require logical reasoning (\textit{and}, \textit{or}). 

\textbf{Traversing reasoning process with a functional program.} With the defined atomic operations, we develop a functional program to traverse the reasoning process by performing the corresponding operations and sequentially updating the explanation based on the collected information. %Inspired by \cite{clevr,gqa}, we develop a functional program to traverse the reasoning process and sequentially update the explanation. 
Inspired by \cite{clevr,gqa}, we represent the reasoning process as a directed graph, where nodes denote the reasoning steps and edges represent their dependencies. As shown in Figure \ref{fig:data}, starting from the initial reasoning step (\textit{i.e.,} \textit{Select (table)}), we recursively construct the partial explanation for the current node (shown to the right of each node in Figure \ref{fig:data}) and pass it to its dependent nodes. Our final explanation is obtained at the last reasoning step (\textit{i.e.,} \textit{And}). To construct the partial explanation for each node, we design a set of templates based on the semantic meanings of the atomic operations (see our supplementary materials for details). The proposed templates dynamically combine the information extracted within the current node and those transited from its dependent nodes in the previous steps. For example, template for the \textit{relate} operation locates a new object based on its relationship with objects selected in the previous nodes.  

% \textit{common} operation takes into account the groups of objects selected in the previous nodes, and find the attributes they have in common.

% For instance, to answer the question in Figure \ref{fig:data}, one needs to locate the table, find the plate on top of it based on their relationship, and investigate the cleanliness and the color of the table. In order to explain the decision-making with joint reasoning on multiple modalities, we explicitly ground the key objects in the corresponding images, instead of modeling visual and textual explanation independently. Finally, our explanation is constructed by executing a functional program that sequentially combines information from different reasoning steps with pre-defined templates, as detailed in the next subsection.

The aforementioned paradigm allows efficiently traversing the reasoning process and constructing explanation that elaborates how a decision is made based on the visual and textual modalities. It not only enables the construction of our new GQA-REX dataset with 1,040,830 multi-modal explanations (data statistics and qualitative examples provided in the supplementary materials), but also plays a key role in improving the interpretability and accuracy of visual reasoning models, as detailed in the next subsection.

\subsection{Explanation Generation Model} \label{exp_generation}
Explaining the rationale behind a decision requires reasoning on visual and textual evidence and elaborating their relationships. Existing explanation generation methods \cite{vqa_x,vqa_e,faithful_exp,transformer_exp} model textual and visual explanations with separate processes, and pay little attention to how key components in each modality correlate with each other. As a result, they have limited capability of generating explanations that jointly consider both modalities and ground words in the images. With the overarching goal of improving the interpretability and accuracy of visual reasoning models, we propose a novel explanation generation model that couples related components across the two modalities and generates the explanation based on their relationships.

\begin{figure}
\centering
\includegraphics[width=0.6\linewidth]{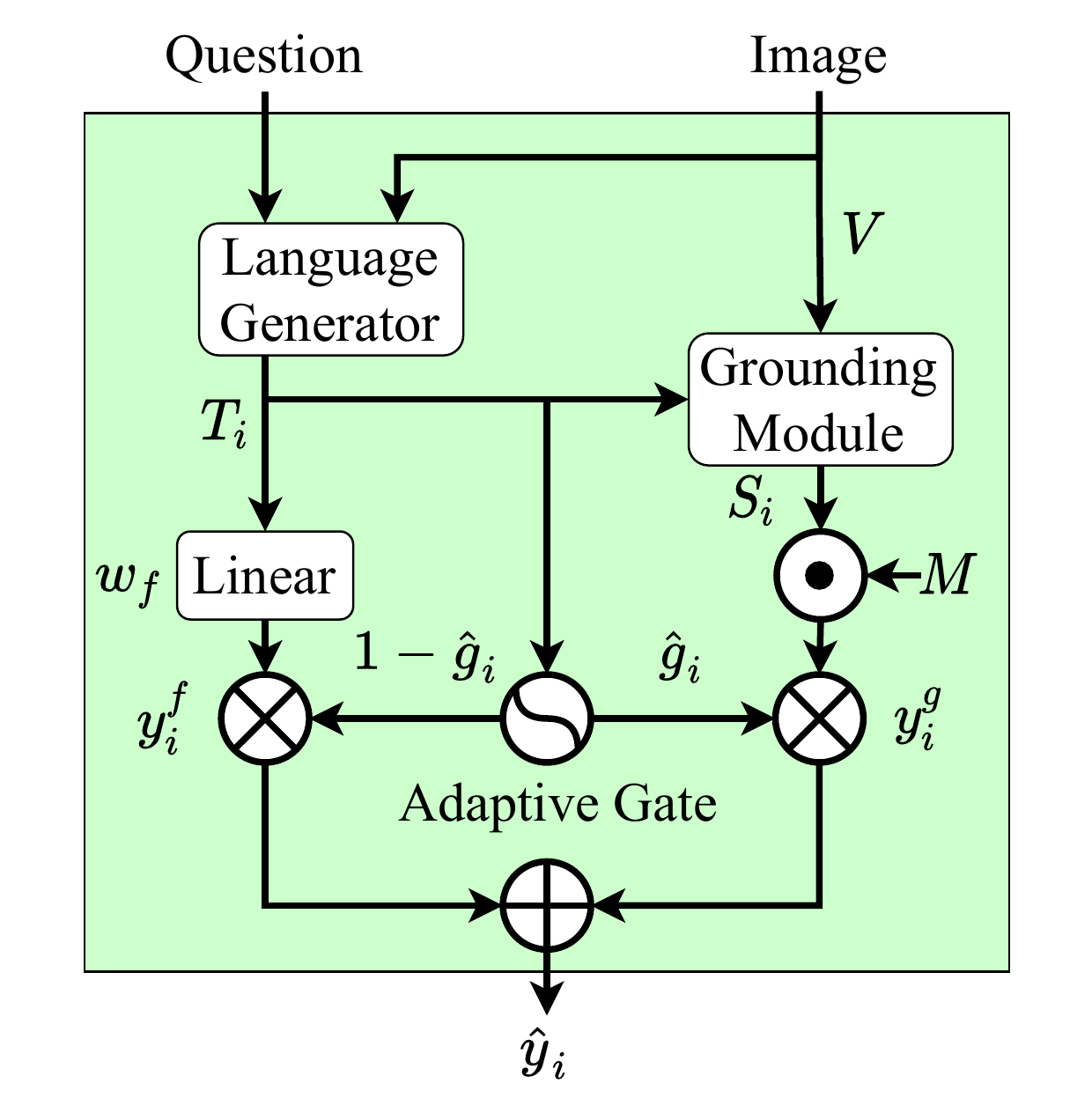}
\caption{Overview of our explanation generation method.}
\label{fig:model}
\end{figure}

Figure \ref{fig:model} illustrates an overview of our method. The principal idea behind the method is to explicitly measure the semantic similarity between words and visual regions, and leverage it to generate multi-modal explanation with enhanced visual grounding. Specifically, unlike conventional methods \cite{vqa_x,vqa_e,faithful_exp,transformer_exp} that generate the explanation solely based on textual features $T_i \in \mathbb{R}^{1 \times D}$ (\textit{e.g.,} LSTM hidden state for predicting the $i^{th}$ word), we further measure the similarity between the textual features $T_{i}$ and visual features $V \in \mathbb{R}^{N \times D}$, and compute the probability of linking the current word with different regions $S_{i} \in \mathbb{R}^{1\times N}$:
\begin{equation}
    S_{i}^{n} = \frac{e^{T_{i} \cdot V_{n}}}{\sum\limits_{j=1}^{N} e^{T_{i} \cdot V_{j}}} 
\end{equation}

\noindent where $N$ denotes the total number of image regions, $D$ is the dimension of features, and $n$ is the index for an image region. $T \cdot V$ is the dot product between two features and corresponds to their cosine similarity.  

To incorporate visual grounding with explanation generation, we leverage a transformation matrix $M \in \mathbb{R}^{N\times K}$ to map grounding results to the prediction of the next word:
\begin{equation}
    y_{i}^{g} = S^{i} \cdot M
\end{equation}
\noindent where $K$ is the number of vocabulary, $M$ is a binary matrix, and $M_{ij} = 1$ if the $j^{th}$ token denotes the $i^{th}$ region (\textit{i.e.,} we use the token $\# i$ to represent grounding a word in the $i^{th}$ region). Since not every word in the explanation can be grounded in the image, \textit{e.g.,} words like ``is'' do not have an associated region, we further develop a gating function to determine if the current word should be grounded:
\begin{equation}
    \hat{g}_{i} = \sigma(W_{g} \cdot T_{i})
\end{equation}
\noindent where $\hat{g}_{i}$ is the probability of grounding the $i^{th}$ word, $W_g \in \mathbb{R}^{1\times D}$ denotes the trainable weights, and $\sigma$ is the sigmoid activation function. We use a balanced binary cross-entropy loss to train the gating function: 
\begin{equation}
    L_{g} = - \sum\limits_{i} \frac{C^{-}}{C} g_{i}\log \hat{g}_{i}+ \frac{C^{+}}{C}(1-g_{i})\log(1-\hat{g}_{i})
\end{equation}
\noindent where $g_{i}$ is the binary ground truth, $C^{+}$ and $C^{-}$ denote the number of grounded and non-grounded words in the current explanation, and $C = C^{+}+C^{-}$. 

Upon obtaining the grounding probability $\hat{g}_{i}$, we adaptively combine the grounding results $y_{i}^{g}$ with the probabilities of different words derived from textual features $y_{i}^{f} = softmax(W_{f} \cdot T_{i})$ to determine the next word $\hat{y}_{i}$:
\begin{equation}
    \hat{y}_{i} = \hat{g}_{i} y_{i}^{g} +(1-\hat{g}_{i}) y_{i}^{f} 
\end{equation}
\noindent where $W \in \mathbb{R}^{K\times D}$ represents trainable weights.

We train our model with a linear combination of the balanced binary cross-entropy loss $L_{g}$ for the gating function and the conventional cross-entropy loss for question answering $L_{ans}$ and explanation generation $L_{exp}$ \cite{vqa_e}:
\begin{equation} \label{eq_loss}
    L = L_{ans}+ L_{exp} + L_{g} 
\end{equation}

With the aforementioned method that couples key components from both modalities, we significantly improve the model's visual grounding capability, which leads to enhanced interpretability and reasoning performance.

\section{Experiments}
In this section, we present the implementation details (Section \ref{implementation}), and conduct experiments to analyze the proposed framework. We first experiment with the conventional multi-task learning paradigm \cite{vqa_e} (Section \ref{main_result}). It demonstrates the effectiveness of our explanation in simultaneously enhancing the models' accuracy and interpretability, and highlights the significance of improving visual grounding with our model. We also conduct experiments under the transfer learning paradigm and analyze different visual skills of the reasoning model, aiming to answer the following research questions:
% \begin{itemize}
%     \item Is the knowledge learned from the explanations transferable to question answering? (Section \ref{transfer_learning})
%     \item How do different visual skills affect answer correctness? (Section \ref{skill_analyse})
% \end{itemize}

(1) Is the knowledge learned from the explanations transferable to question answering? (Section \ref{transfer_learning})

(2) How do different visual skills affect answer correctness? (Section \ref{skill_analyse})

\subsection{Implementations} \label{implementation}
\begin{table*}
\centering
\resizebox{1\textwidth}{!}{
% {\smal
\begin{tabular}{c |c c c c c |c | c c c c}
\toprule
 & BLEU-4 & METEOR & ROUGE-L & CIDEr & SPICE & Grounding & GQA-val & GQA-test & OOD-val & OOD-test \\
% \midrule
\hline
VisualBert \cite{visualbert} & - & - & - & - & - & - & 64.14 & 56.41 & 48.70 & 47.03 \\

VisualBert-VQAE \cite{vqa_e} & 42.56 & 34.51 & 73.59 & 358.20 & 40.39 & 31.29 & 65.19 & 57.24 & 49.20 & 46.28 \\

VisualBert-EXP \cite{faithful_exp} & 42.45 & 34.46 & 73.51 & 357.10 & 40.35 & 33.52 & 65.17 & 56.92 & 49.43 & 47.69 \\

VisualBert-REX & \textbf{54.59} & \textbf{39.22} & \textbf{78.56} & \textbf{464.20} & \textbf{46.80} & \textbf{67.95} & \textbf{66.16} & \textbf{57.77} & \textbf{50.26} & \textbf{48.26}  \\  
\bottomrule
\end{tabular}
}
\caption{Comparative results on explanation generation and question answering. GQA- and OOD- denote results on GQA and GQA-OOD. Best results are highlighted in bold.}
\label{multi_task_learning}
\end{table*} 

\textbf{Dataset.} We experiment with our proposed GQA-REX dataset, which is constructed based on the balanced training and validation sets of GQA \cite{gqa}. We optimize models on the training set and evaluate their performance of explanation generation on the validation set. To evaluate the reasoning performance, we adopt the balanced validation and test-standard sets of GQA. %The training and validation set have 943000 and 132062 questions. The test-dev set of GQA has 12578 questions but does not provide scene graph annotations for generating our explanation, thus we only use it to evaluate the reasoning performance. 
Since the annotated bounding boxes for visual grounding may not align with the visual inputs (\textit{i.e.}, UpDown regional features \cite{updown}), we convert the grounding annotations into a set of tokens by finding the input region that has the highest Intersection of Union (IoU) with the ground truth bounding box (\textit{i.e.,} $\#i$ means the bounding box is aligned with the $i^{th}$ input region). We also experiment with the recently introduced GQA-OOD dataset \cite{gqa_ood} with out-of-distribution data (\textit{i.e.,} ``tail'' questions).

\textbf{Evaluation.} We evaluate models from multiple perspectives, including reasoning performance, quality of explanations, visual grounding, and recognition of attributes. We use answer accuracy for evaluating reasoning performance. For the quality of explanations, we follow \cite{vqa_x,faithful_exp,vqa_e} and adopt five language evaluation metrics, including BLEU-4 \cite{bleu}, METEOR \cite{meteor}, ROUGE-L \cite{rouge}, CIDEr \cite{cider}, and SPICE \cite{spice}. Similar to \cite{gqa}, visual grounding (\textit{i.e.}, Grounding) is evaluated by aggregating grounded regions in the predicted explanations and computing their IoU with the ground truth. We evaluate the recognition of eight unique types of attributes, including color, material, sport, shape, pose, size, activity, and relation. We only consider samples where the attributes do not appear in the questions to avoid trivial solutions, and calculate the recall of predicting the correct attributes in the explanations.

\textbf{Model specification.} We use the state-of-the-art VisualBert \cite{visualbert} with UpDown regional features \cite{updown} as the backbone for visual reasoning (more details in the supplementary materials). The model is pretrained on MSCOCO \cite{mscoco} dataset without using annotations for question answering. Therefore, it enables us to study the transferability of knowledge learned from our explanations. For explanation generation, we adopt the language generator developed in \cite{faithful_exp} as our baseline, and incorporate our method proposed in Section \ref{exp_generation} to enhance its grounding capability.

\textbf{Training.} We use Adam \cite{adam} optimizer to train the models with batch size 128. For multi-task learning paradigm \cite{vqa_e}, we train the model to simultaneously predict answers and explanations for 8 epochs. The learning rate is initialized as $10^{-4}$ and decayed once by $0.25$ at the last epoch. For the transfer learning paradigm (Section \ref{transfer_learning}), we first train the models on explanation generation for 8 epochs and then fine-tune them under the multi-task learning paradigm for 15 epochs. The learning rate is decayed at the $8^{th}$ and $12^{th}$ epoch, respectively.

\begin{figure*}
\centering
\includegraphics[width=0.95\linewidth]{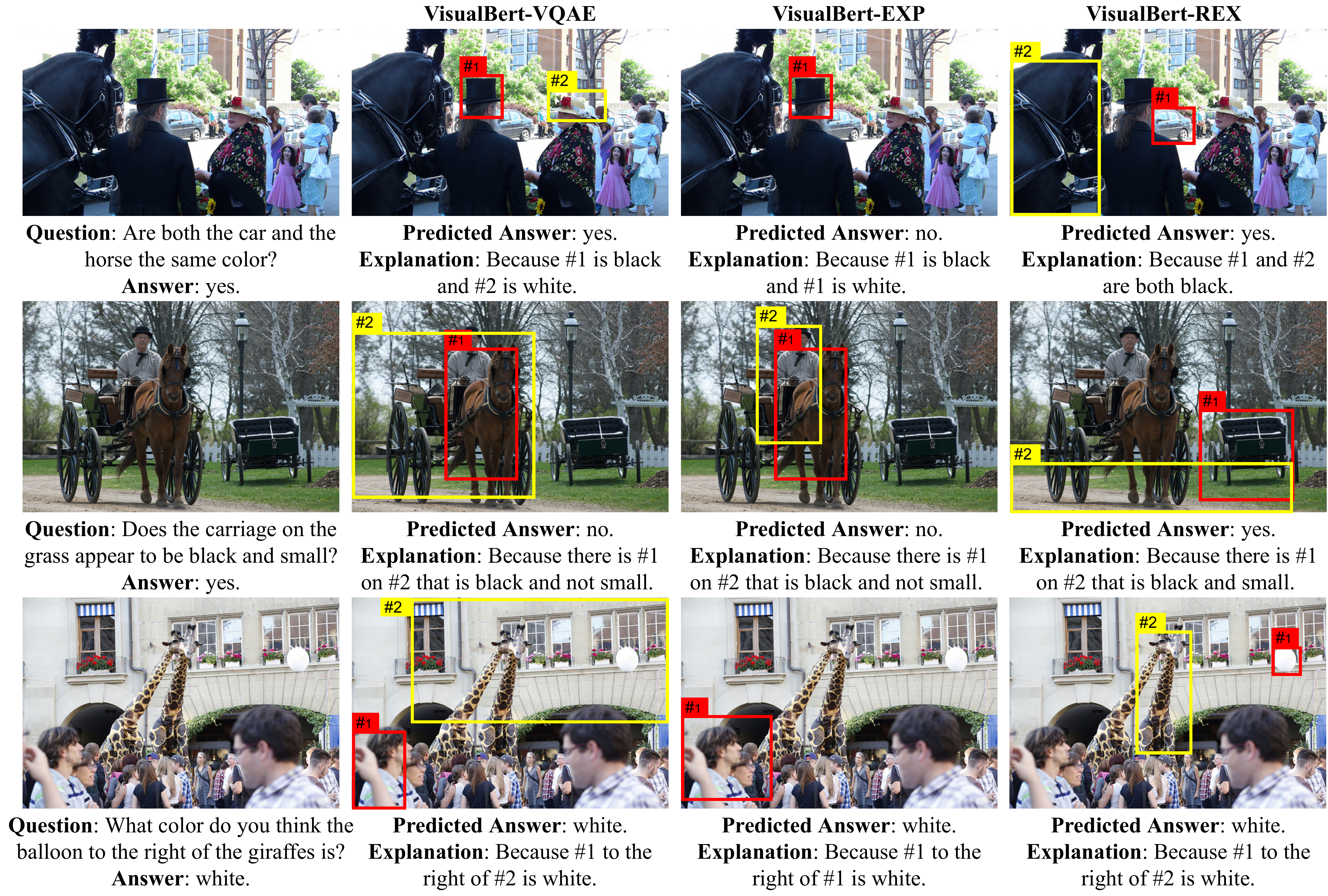}
\caption{Qualitative results for explaining models' decision-making process. Visual grounding is represented with the token $\#$.}
\label{fig:qualitative}
\end{figure*}

\subsection{Results} \label{main_result}
We first validate the effectiveness of our framework under the multi-task learning paradigm \cite{vqa_e}. We compared our model (\textit{i.e.,} VisualBert-REX) with three approaches with the same backbone, including the VQA baseline (\textit{i.e.,}  VisualBert \cite{visualbert}), and two explanation generation methods (\textit{i.e.,} VisualBert-VQAE \cite{vqa_e} and VisualBert-EXP \cite{faithful_exp}).

As shown in Table \ref{multi_task_learning}, learning with both answers and explanations (\textit{i.e.,} VisualBert-VQAE and VisualBert-EXP) leads to a reasonable improvement over the counterpart using answers alone (\textit{i.e.,} VisualBert), and can further provide illustration of the decision-making process. It shows that our proposed explanations can complement the answer annotations and simultaneously increase the accuracy and interpretability of visual reasoning models. However, it is important to note that, the existing approaches lack the capability of correlating words with their corresponding regions of interest, and thus have low visual grounding scores. Differently, by explicitly modeling the correspondence between key components across the visual and textual modalities, our VisualBert-REX method significantly increases the visual grounding score, leading to further improvements in the quality of the generated explanations and reasoning performance. These observations validate the usefulness of our explanations, and highlight the advantages of our explanation generation method in augmenting visual reasoning models with enhanced visual grounding capability.

In addition to the quantitative evaluation, we also perform qualitative analyses on the predicted answers and explanations. As shown in Figure \ref{fig:qualitative}, our VisualBert-REX explains the rationales behind decisions with high-quality explanations, which leads to more accurate answers. Unlike VisualBert-VQAE and VisualBert-EXP that have difficulties localizing the important regions (\textit{e.g.,} horse and car in the $1^{st}$ example), it accurately captures the key objects with enhanced visual grounding capability and compares their attributes to answer correctly. It also avoids hard-negative objects (\textit{e.g.,} the large carriage on the road in the $2^{nd}$ sample) by investigating the relationships between objects (\textit{e.g.,} carriage on the grass). Moreover, while conventional methods generate answers without actually reasoning on the visual observations (\textit{e.g.,} without focusing on key objects in the $3^{rd}$ sample), our method faithfully answers the questions by reasoning on all regions of interest. 

\begin{table*}
\centering
\resizebox{0.7\linewidth}{!}{
{\small
\begin{tabular}{cccccccc}
\toprule
\multicolumn{2}{c}{\multirow{2}{*}{}} & \multicolumn{2}{c}{1$\%$} & \multicolumn{2}{c}{5$\%$} & \multicolumn{2}{c}{10$\%$} \\
\cmidrule(lr){3-4} \cmidrule(lr){5-6} \cmidrule(lr){7-8}
\multicolumn{2}{c}{} & GQA & OOD & GQA & OOD & GQA & OOD \\ 
\midrule
VQA-only & VisualBert & 41.41 & 27.11 & 48.53 & 33.78 & 51.79 & 37.83 \\ 
\midrule
\multirow{2}{*}{Multi-task learning} & VisualBert-EXP & 41.70 & 27.09 & 49.36 & 34.50 & 52.83 & 38.33 \\ 
& VisualBert-REX & 40.42 & 23.95 & 50.30 & 35.69 & 53.90 & 40.08 \\ 
\midrule
Self-supervised learning & VisualBert & 45.06 & 30.62 & 52.12 & 38.68 & 54.74 & 40.12 \\ 
\midrule
\multirow{2}{*}{Transfer learning} & VisualBert-EXP & 51.32 & 35.26 & 56.34 & 41.20 & 57.65 & 43.15 \\ 
& VisualBert-REX & \textbf{57.07} & \textbf{40.03} & \textbf{61.28} & \textbf{45.02} & \textbf{61.90} & \textbf{45.98} \\ 
\bottomrule
\end{tabular}
}
}
\caption{Comparative results for models trained using different proportions of answer annotations. Results are reported on the balanced validation set of GQA and the validation set of GQA-OOD. Best results are highlighted in bold.}
\label{transfer_learning_res}
\end{table*} 

\subsection{Is the knowledge learned from the explanations transferable to question answering?} \label{transfer_learning}
Previous methods either simultaneously answer questions and generate explanations \cite{vqa_e} or generate explanations for fixed answers \cite{vqa_x,faithful_exp,competing_exp}, and pay little attention to how transferable is knowledge learned from explanations. Inspired by the recent study \cite{frozen} that shows knowledge learned from text corpus can enable few-shot visual question answering, in this section, we evaluate the transferability of the proposed reasoning-aware and grounded explanation and analyze its usefulness in distilling knowledge from the reasoning process into question answering.

Specifically, we consider a transfer learning paradigm: first training the models on explanation generation with our complete training set, and then fine-tuning them on both explanation generation and question answering with a subset of 1$\%$, 5$\%$, and 10$\%$ of training data. The subsets are created by randomly sampling the specific proportions of questions from each reasoning type, so that the overall statistics about different reasoning tasks are well preserved. We evaluate models on the complete validation set regardless of the amount of training data. To demonstrate the transferability of the knowledge learned from our explanations, we compare the aforementioned method with three alternatives: We first consider (1) a VQA-only and (2) a multi-task learning baseline. They are identical to those discussed in Section \ref{main_result} but trained only on the respective subsets, and thus do not benefit from the knowledge transferred from explanations. (3) To validate that the improvements achieved by transfer learning come from the explanations instead of access to additional questions, we further compare with a self-supervised learning method that pretrains the model on all training questions under the BERT\cite{bert} paradigm and then fine-tunes it on the subsets for question answering. Two observations can be made on the results reported in Table \ref{transfer_learning_res}:  
% \begin{itemize}
%     \item \textbf{Knowledge transferred from the explanations plays a key role in question answering.} By incorporating the proposed explanation data, models trained under the multi-task learning paradigm outperform the VQA-only baseline despite the scarcity of data. Moreover, with more abundant knowledge transferred from the explanations, the transfer learning method improves the performance by a large margin and achieves the best results regardless of the amount of answer annotations. It is notable that VisualBert-REX with only 10$\%$ of answer annotations can achieve comparable performance to VisualBert trained on the complete training set. On the contrary, while self-supervised learning also increases the reasoning performance, it is not as effective as the transfer learning method. These observations demonstrate the transferability of knowledge learned from our explanations, and highlight its role in establishing understanding of the reasoning process for more efficient learning on visual reasoning.
%     \item \textbf{Visual grounding is important for transferring knowledge.} Compared to VisualBert-EXP, VisualBert-REX with enhanced visual grounding capability achieves much better results under the transfer learning paradigm, demonstrating the advantages of our method under various training paradigms. More importantly, it highlights the significance of visual grounding for developing better understanding of the reasoning process and transferring knowledge from explanations to question answering.    
% \end{itemize}

\textbf{Knowledge transferred from the explanations plays a key role in question answering.} By incorporating our explanations, models trained under the multi-task learning paradigm outperform the VQA-only baseline despite the scarcity of data. Moreover, with more abundant knowledge transferred from the explanations, the transfer learning method improves the performance by a large margin and achieves the best results regardless of the amount of answer annotations. It is notable that VisualBert-REX with only 10$\%$ of answers achieves comparable performance to VisualBert trained on the complete training set. On the contrary, while self-supervised learning also increases the reasoning performance, it is not as effective as transfer learning. These observations demonstrate the transferability of knowledge learned from our explanations, and highlight its role in establishing understanding of the reasoning process for more efficient learning on visual reasoning.

\textbf{Visual grounding is important for transferring knowledge.} Compared to VisualBert-EXP, VisualBert-REX with enhanced visual grounding capability achieves much better results under the transfer learning paradigm, demonstrating the advantages of our method under various training paradigms. More importantly, it highlights the significance of visual grounding for developing better understanding of the reasoning process and transferring knowledge from explanations to question answering.    

\subsection{How do different visual skills affect answer correctness?} \label{skill_analyse}
Answering visual questions involves performing various visual skills \cite{visual_skill}, \textit{e.g.,} recognition of objects' attributes such as colors and positional relationships. Existing studies assess these skills by categorizing questions into different groups and analyzing them separately \cite{vqa1,vqa2,gqa,visual_skill}. While showing usefulness in studying the models' capability of tackling different types of questions, they fall short of explaining the relationship between successfully performing a skill and correctly answering the question. In this paper, we use a more explicit approach to analyze how various skills affect the answer correctness. Specifically, we evaluate the recognition of eight common attributes based on the model's capability to derive the corresponding concepts in the explanations, \textit{e.g.,} a successful recognition of color requires the model to explain its decision with the key colors, and leverage recall rates of capturing the concepts for quantitative analyses. %For visual grounding, we adopt the grounding score discussed in Section \ref{implementation}. 
In Table \ref{skill_correlation}, we report the evaluation scores of different skills and their Pearson's correlation with the predicted probabilities on the correct answers. We make two observations on the results:

\begin{table}
\centering
\resizebox{0.9\linewidth}{!}{
{\small
\begin{tabular}{ccc}
\toprule
 & Recall Rate & Pearson's Correlation\\
\midrule
Color & 56.01 & 0.742 \\
\midrule
Material & 49.27 & 0.708 \\
\midrule
Sport & 72.77 & 0.575 \\
\midrule
Shape & 40.64 & 0.548 \\
\midrule
Pose & 74.80 & 0.417 \\
\midrule
Size & 65.31 & 0.574 \\
\midrule
Activity & 46.58 & 0.666 \\
\midrule
Relation & 29.00 & 0.182 \\
% \midrule 
% Visual grounding & 67.95 & -0.002 \\
\bottomrule
\end{tabular}
}
}
\caption{Recall rates for capturing key concepts related to different visual skills, and their correlation with the reasoning performance.}
\label{skill_correlation}
\end{table} 

\textbf{Recognition of attributes is important for answering correctly.} Our results show that the recall rates for all attributes have reasonable correlation with the reasoning performance, which validates the importance of capturing key attributes in the images for answering visual questions. 

\textbf{Attributes do not contribute equally to answer correctness.} It is notable that the model has diverse performance on recognizing different attributes, and the differences in the correlation between skills and answer correctness are also significant (\textit{e.g.}, recognition of colors v.s. recognition of relations). The results indicate that, while it is important for humans to capture different key attributes in order to answer correctly, the attributes do not contribute equally to the decision making of a computational model.

Our results shed light on the underlying decision making of visual reasoning models, and reveal the influences of various visual skills on answer correctness.

\section{Discussion}
We introduce REX, a principled framework with a new type of reasoning-aware and grounded explanations, a functional program for automatically constructing explanations, and a novel explanation generation method that explicitly couples key components in different modalities. Experimental results demonstrate the usefulness of our framework in explaining the models' decision-making process and improving the visual reasoning performance. They also highlight the critical need to increase models' visual grounding capability for understanding the reasoning process. 

\textbf{Limitations.} Despite the aforementioned advantages of our data and model, we believe there is still a large room for interpretable visual reasoning. While the proposed data offers multi-modal explanations derived from a diverse set of images and vocabulary, it may still fail to cover all types of real-world problems. For example, some questions may require external knowledge that is unavailable in the given visual-textual data \cite{okvqa}. One possible direction to address the challenges is to incorporate explanations with vision-and-language pretraining on external knowledge base (\textit{e.g.,} \cite{conceptnet}), as our experiments show the effectiveness of transferring knowledge with explanations.  

\section{Broader Impact}
Endowing AI systems the ability to elaborate their decision-making process with high quality multi-modal explanations is an important step toward trustworthy AI. It could fundamentally address the critical needs in opening the black-box of AI algorithms. We therefore envision this work to offer new opportunities to a wide range of domains especially those taking interpretability and transparency as a high priority, such as healthcare, finance, and legislation. The new paradigm highlighting multi-modal understanding, and the large-scale dataset with diverse high-quality explanations may spur innovations and developments in these areas. The explanation generation model elucidates the reasoning process and key components in decision making, alleviating safety or fairness risks of decision-critical applications. We hope that this work would be a useful resource and open a new avenue for the community to develop interpretable and transparent AI systems.

% We envision that future artificial intelligence (AI) systems will not only assist users in addressing different problems but also enable the understanding of how they make decisions. We believe that our proposed data and model can be useful for building trust between general users and the AI systems. It will also offer the opportunities to users for reconsidering their decisions based on the explanations of the models and adjust them accordingly. 

% We do not foresee negative ethical consequences as a direct result of this work. However, our work centers around the visual reasoning tasks, where existing models are prone to data biases. The answers provided by the models could be biased and do not accurately reflect the visual world, which can cause misunderstanding among users (\textit{e.g.}, those who are visually impaired and have higher reliance on AI systems). We hope that our work could be useful for the development of visual reasoning models that are reliable in real-world scenarios.  

\section*{Acknowledgements}
This work is supported by NSF Grants 1908711 and 1849107.

%%%%%%%%% REFERENCES
{\small
\bibliographystyle{ieee_fullname}
\bibliography{egbib}
}

\clearpage

%%%%%%%%% Supplementary 
\section{Supplementary Materials}
This supplementary materials provide additional details of the proposed framework. Specifically,
\begin{itemize}
    \item We elaborate the detailed templates for constructing partial explanations with atomic operations (Section \ref{template}). 
    \item We present the statistics of the proposed GQA-REX dataset, and provide qualitative examples of our reasoning-aware and grounded explanations (Section \ref{dataset}).
    \item We provide the implementation details of our proposed explanation generation method (Section \ref{implementation}).
\end{itemize}

\subsection{Templates for Constructing Explanations} \label{template}
Our functional program proposed in the main paper progressively traverses the reasoning process and uses pre-defined templates to construct partial explanation at each reasoning step. In this section, we present the details of the templates. Our templates are designed based on the semantic meaning of each atomic operation, and take into account both information extracted in the current reasoning step and that passed from previous steps. As shown in Table \ref{atomic_operation}, we define three general functions shared across different templates: [OBJ] selects a specific type of visually grounded objects, [ATTR] finds desired attributes specified in the atomic operation, and [DEP] collects partial explanations from dependent nodes in the previous steps. Other functions are more specific to a single atomic operation: [CHECK\_EXISTENCE] examines if a certain type of objects exist in the scene; [QUERY\_ATTR] queries the value of a specific type of attributes; [VERIFY\_ATTR] examines if the selected objects have certain attributes; [FIND\_COMMON] finds the commons attributes shared by both groups of objects; [COMPARE\_ATTR] compares two groups of objects based on a specific type of attributes; [RELATION] finds the desired relationships between two groups of objects; [LOGICAL AND/OR] denotes logical operations.
\begin{table}[t]
\begin{center}
\resizebox{0.49\textwidth}{!}{
\begin{tabular}{cc}
\toprule
\textbf{Operation} & \textbf{Template}  \\
\midrule
Select & [OBJ] \\
\midrule
Exist & There [CHECK\_EXISTENCE] [DEP] \\
\midrule
Filter & [ATTR] [OBJ]  \\
\midrule
Query  & [DEP] is/are [QUERY\_ATTR]  \\
\midrule
Verify & [DEP] is/are [VERIFY\_ATTR] \\
\midrule
Common & both [DEP 1] and [DEP 2] are [FIND\_COMMON] \\
\midrule
Same  & [DEP 1][DEP 2] are [ATTR] \ / \ [DEP1] is [ATTR1] and [DEP 2] is [ATTR 2] \\
\midrule
Different & [DEP1] is [ATTR1] and [DEP 2] is [ATTR 2] \ / \ [DEP 1][DEP 2] are [ATTR] \\
\midrule 
Compare & [DEP 1] is [COMPARE\_ATTR] than [DEP 2]  \\
\midrule
Relate & [DEP 1] [RELATION] [DEP 2] \\
\midrule
And/Or & [DEP 1] [LOGICAL AND/OR] [DEP 2] \\
\bottomrule
\end{tabular}
}
\end{center}
\caption{Templates for constructing partial explanation with atomic operations.}
\label{atomic_operation}
\end{table}

With the aforementioned templates, we sequentially update the explanation by selectively attending to different regions of interest, investigating the desired attributes, and accumulating information along the reasoning-process. The templates not only enable the collection of our new GQA-REX dataset, but also provide a general paradigm for automatically constructing explanations based on the reasoning process.

\subsection{The GQA-REX Dataset} \label{dataset}
Aiming to provide an explanation benchmark that encodes the reasoning process and grounding across the visual-textual modalities, we propose a new GQA-REX dataset that consists of 1,040,830 reasoning-aware and visually grounding explanations. In this section, we present the statistics of the our dataset, including the distribution of atomic operations and the distribution of visually grounding objects. We also provide qualitative examples of our defined explanations. 

%They are grounded on the corresponding functional program in GQA \cite{gqa} for generating the question, and makes use of the ground truth scene graph for localizing regions of interest. 

\begin{figure}
\centering
\includegraphics[width=1\linewidth]{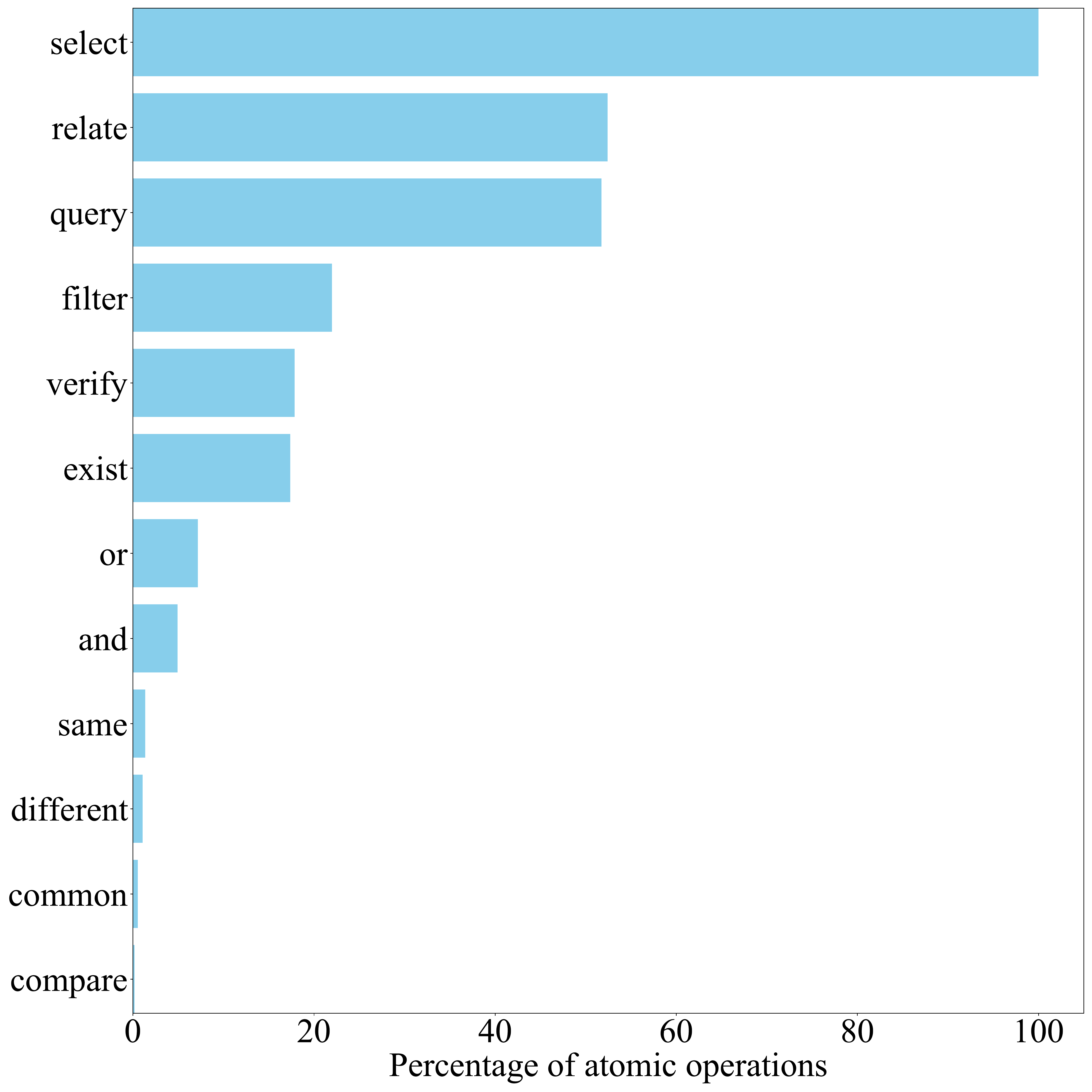}
\caption{Percentage of questions with different types of reasoning operations.}
\label{supp:operation}
\end{figure}

\begin{figure}
\centering
\includegraphics[width=1\linewidth]{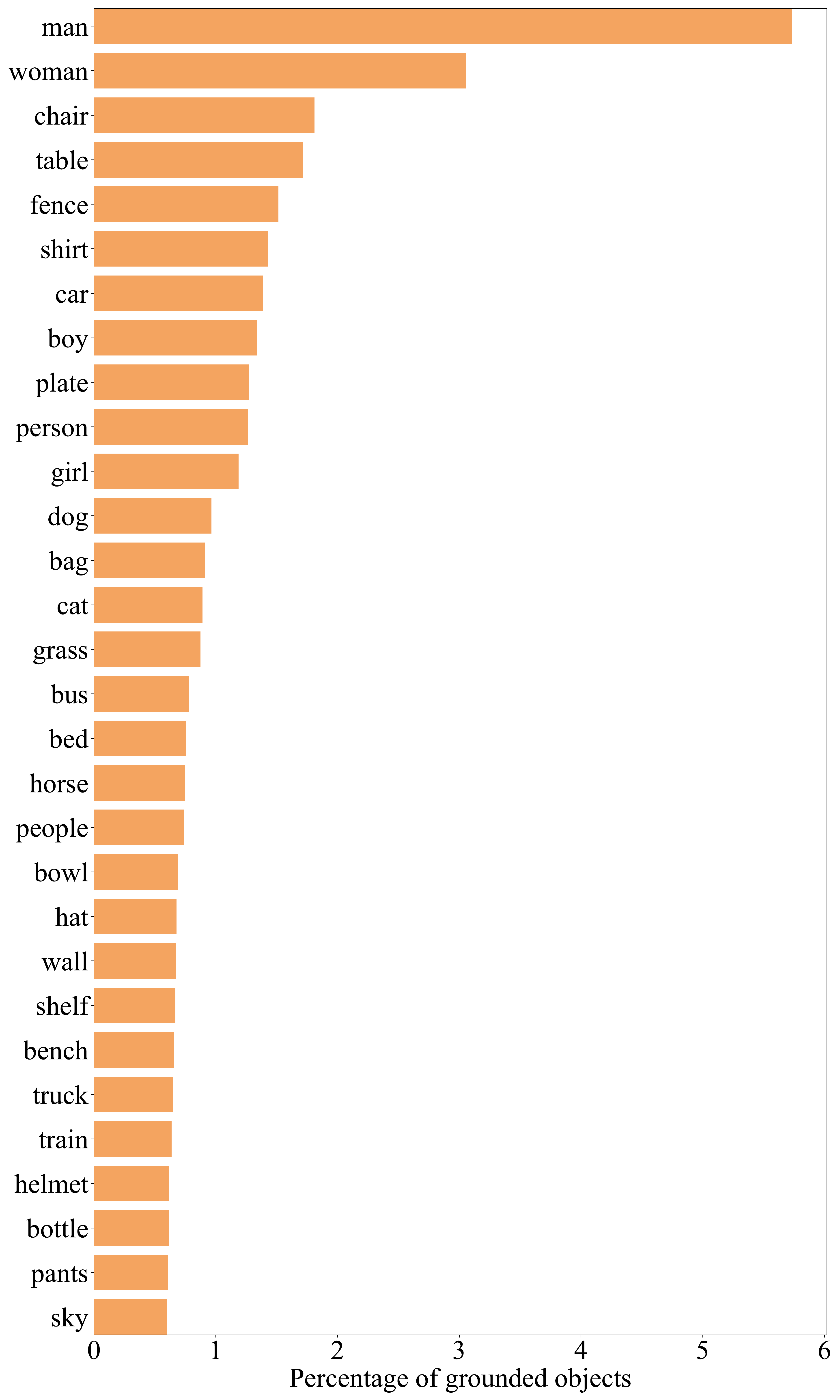}
\caption{Top-20 object categories for visual grounding and their percentages in our dataset.}
\label{supp:grounding}
\end{figure}

As shown in Figure \ref{supp:operation}, our dataset covers the explanations for a wide range of visual questions: All of the questions require attending to specific regions of interest (\textit{i.e.,} the \textit{select} operation) to derive the answers, which highlights the need to explain decisions with visual grounding. A large proportion of questions involve recognizing certain attributes (\textit{i.e.,} \textit{relate}, \textit{query}, \textit{filter}, and \textit{verify}), which is one of the fundamental skills for understanding the visual world. Some questions require examining the existence of certain types of objects or performing logical operations, which correspond to the considerable amount of yes/no questions. There are also relatively difficult questions that ask models to investigate all attributes of two groups of objects (\textit{i.e.,} \textit{same}, \textit{different}, \textit{common}, and \textit{compare}). 

To explain the decision making for various questions with multi-modal evidence, we link a diverse collection of objects with their corresponding regions of interest. As shown in Figure \ref{supp:grounding}, unlike \cite{vcr} that focuses on grounding a single type of objects (\textit{i.e.,} humans), our dataset takes into account 1,660 unique types of object categories and provides more fine-grained categorization (\textit{e.g.,} human characters are categorized based on their genders, \textit{i.e.,} woman and man, and ages, \textit{i.e.,} boy and man). The visual grounding plays an essential role in explaining how different components in the visual-textual modalities contribute to the decision-making, and enabling the development of computational models with improved multi-modal understanding of the reasoning process (\textit{e.g.,} VisualBert-REX in the main paper).

\begin{figure*}
\centering
\includegraphics[width=0.9\linewidth]{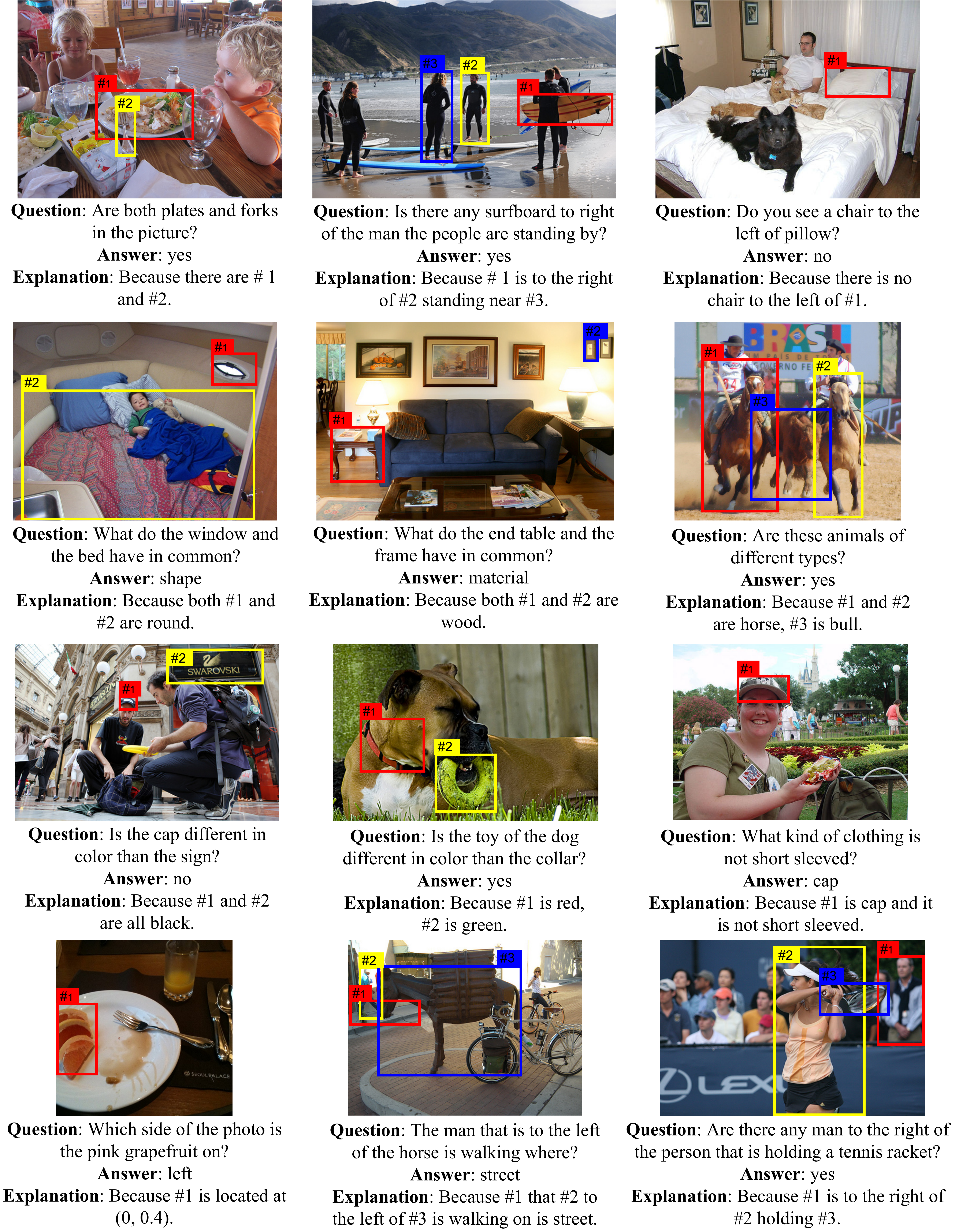}
\caption{Qualitative examples of explanations in our GQA-REX dataset.}
\label{supp:example}
\end{figure*}

In Figure \ref{supp:example}, we visualize examples of our explanations for different types of questions, \textit{e.g.}, questions examining the existence of certain object in the $1^{st}$ row, questions relating to the attributes of multiple objects in the $2^{nd}$ and the $3^{rd}$ rows, and questions investigating different types of relationships in the $4^{th}$ row. They demonstrate the effectiveness of our defined explanations on elaborating the rationales behind the answers, and validate the usefulness of our functional program in automatically constructing the explanations.

\subsection{Implementation of VisualBert-REX} \label{implementation}
In this section, we provide the implementation details of the proposed explanation generation method, \textit{i.e.,} VisualBert-REX in the main paper. Similar to the VisualBert-EXP baseline, our method adopts the state-of-the-art VisualBert \cite{visualbert} as our visual reasoning backbone and the LSTM-based language generator from \cite{faithful_exp}, and jointly predicts the answer and corresponding explanation. We concatenate word embeddings of the question and UpDown regional features \cite{updown}, and use VisualBert to learn cross-modal features from them. Cross-modal features extracted at first token (\textit{i.e.,} [CLS]) is utilized for predicting the answer and initializing the hidden state of the language generator. When sequentially generating each word in the explanation, we measure the similarity between the hidden state for the current step and the VisualBert features for all visual regions, and normalize the results to obtain the probabilities of grounding the current word in specific regions (\textit{i.e.,} $y_{i}^{g}$ in Equation 5 of the main paper). The grounding result is adaptively combined with the prediction determined based on the hidden state (\textit{i.e.,} $y_{i}^{f}$ in Equation 5 of the main paper), and the combined result is used to determine the next word in the explanation.

% {\small
% \bibliographystyle{ieee_fullname}
% \bibliography{egbib}
% }

\end{document}